%% file: main.tex
\documentclass[conference]{IEEEtran}
\IEEEoverridecommandlockouts
\usepackage{cite}
\usepackage{amsmath,amssymb,amsfonts}
\usepackage{algorithmic}
\usepackage{graphicx}
\usepackage{textcomp}
\usepackage{xcolor}
\usepackage{subfiles}

\usepackage{makecell}

\usepackage{url}
\usepackage{scrextend}

\usepackage [english]{babel}
\usepackage [autostyle, english = american]{csquotes}
\MakeOuterQuote{"}

\def\BibTeX{{\rm B\kern-.05em{\sc i\kern-.025em b}\kern-.08em
    T\kern-.1667em\lower.7ex\hbox{E}\kern-.125emX}}
\begin{document}

\title{The Scope of In-Context Learning for the Extraction of Medical Temporal Constraints}

\author{\IEEEauthorblockN{Parker Seegmiller}
\IEEEauthorblockA{\textit{Dartmouth College}\\
matthew.p.seegmiller.gr@dartmouth.edu}
\\
\IEEEauthorblockN{Diane Cook}
\IEEEauthorblockA{\textit{Washington State University}\\
djcook@wsu.edu}

\and
\IEEEauthorblockN{Joseph Gatto}
\IEEEauthorblockA{\textit{Dartmouth College}\\
Joseph.M.Gatto.GR@dartmouth.edu}
\\
\IEEEauthorblockN{Hassan Ghasemzadeh}
\IEEEauthorblockA{\textit{Arizona State University}\\
Hassan.Ghasemzadeh@asu.edu}
\\
\IEEEauthorblockN{Sarah Preum}
\IEEEauthorblockA{\textit{Dartmouth College}\\
Sarah.Masud.Preum@dartmouth.edu}

\and
\IEEEauthorblockN{Madhusudan Basak}
\IEEEauthorblockA{\textit{Dartmouth College}\\
Madhusudan.Basak.GR@dartmouth.edu}
\\
\IEEEauthorblockN{John Stankovic}
\IEEEauthorblockA{\textit{University of Virginia}\\
stankovic@cs.virginia.edu}
}

\maketitle

\begin{abstract}
Medications often impose temporal constraints on everyday patient activity. Violations of such medical temporal constraints (MTCs) lead to a lack of treatment adherence, in addition to poor health outcomes and increased healthcare expenses. These MTCs are found in drug usage guidelines (DUGs) in both patient education materials and clinical texts. Computationally representing MTCs in DUGs will advance patient-centric healthcare applications by helping to define safe patient activity patterns. We define a novel taxonomy of MTCs found in DUGs and develop a novel context-free grammar (CFG) to computationally represent MTCs from unstructured DUGs. Additionally, we release three new datasets with a combined total of $N=836$ DUGs labeled with normalized MTCs. We develop an in-context learning (ICL) solution for automatically extracting and normalizing MTCs found in DUGs, achieving an average F1 score of $0.62$ across all datasets. Finally, we rigorously investigate ICL model performance against a baseline model, across datasets and MTC types, and through in-depth error analysis.
\end{abstract}


\subfile{sections/introduction}
\subfile{sections/solution}
\subfile{sections/data}
\subfile{sections/experiments}

\subfile{sections/relatedworks}
\subfile{sections/conclusion}

\bibliography{mtcbib}{}
\bibliographystyle{plain}

\end{document}

%% file: sections/introduction.tex
\section{Introduction}

According to the CDC, over 48\% of the US population uses at least one prescription medicine, and 24\% take three or more \cite{CDCPrescription}. However, only four out of every five new prescriptions are filled, and half of those are administered inappropriately \cite{osterberg2005adherence}. Non-adherence includes incorrectly taking medication concerning a prescription's suggested time, dosage, frequency, or duration. Non-adherence also includes the mistiming of medication intake with respect to other activities when medication efficacy is temporally dependent on those activities, e.g. eating, exercising, or sleeping \cite{osterberg2005adherence, ingersoll2008impact, ferguson2017barriers, klakegg2018assisted}. We refer to any temporal constraints associated with medications as medical temporal constraints (MTCs). Non-adherence to MTCs is linked to higher hospital admission rates, increased morbidity, higher healthcare expenses, poor health outcomes, and even death \cite{dimatteo2004variations, chisholm2012cost, CDCPrescription2, ferguson2017barriers, klakegg2018assisted}. The effect of violating MTCs can range from minor discomfort to emergency room visits \cite{pham2011national}.

MTCs are found in drug usage guidelines (DUGs), or medication guidelines. These textual guidelines appear in both formal patient education materials (e.g., drug labels or public health websites \cite{openFDA, Medscape}), as well as in clinical texts (e.g., prescriptions and after-visit summaries recorded in electronic health records (EHRs) \cite{mtsamples}). The variety of MTC sources calls for a generalizable approach to extract and normalize MTCs from such heterogeneous sources. 

Although MTCs are critical for medical safety and treatment adherence, to our knowledge, there is no existing solution to formulate and model patient-centric MTCs. This requires (i) creating a flexible and robust computational representation of MTCs, (ii) a dataset of natural language descriptions of MTCs annotated with their computational representations, and (iii) a generalizable solution for mapping descriptions of MTCs to their corresponding computational representations. Addressing these challenges can enhance intelligent systems that improve medication adherence and patient safety \cite{roca2021validation, klakegg2018assisted, preum2021review, stankovic2021challenges} or text-based solutions to recommend safe, personalized health information \cite{preum2017preclude2, preclude}.

We formulate and model MTCs for treatment adherence and health safety, in addition to benchmarking the task of extracting MTCs from DUGs. Specifically, (i) we develop a novel taxonomy of potential MTCs and a novel context-free grammar (CFG) based model to represent MTCs from unstructured DUGs computationally. Next, (ii) the taxonomy and CFG are used to label MTCs in three datasets of free-format textual DUGs from heterogeneous sources. Finally, (iii) we define and benchmark the MTC extraction and normalization task using state-of-the-art in-context learning (ICL) strategies, achieving an average F1 score of $0.62$ across all datasets. Recent work has demonstrated the generalizability of ICL for extracting health information in the few-shot setting \cite{agrawal2022large, dunn2022structured, torii2023task}. ICL utilizes a large language model (LLM) to perform a task by conditioning on a few input-output examples. We also compare ICL to a rule-based baseline model, explore several prompting techniques for ICL, and conduct a thorough error analysis to determine the scope of ICL for this new, safety-critical medical NLP task. 




%% file: sections/solution.tex
\section{Medical Temporal Constraints (MTCs)}

Modeling MTCs in DUGs is challenging for the following reasons. MTCs vary in terms of temporal precision; some MTCs are definitive, while some are imprecise. Many MTCs constrain a single activity, the medication intake activity (e.g., taking a medication at \textit{n} hour intervals). However, MTCs can also form dependencies between multiple activities (e.g., taking a medication \textit{m} hours before eating). Based on our review of DUGs from heterogeneous sources \cite{preum2018corpus, Medscape, openFDA, mtsamples}, we formulate the following novel taxonomy of MTCs.

\subsection{Taxonomy of MTCs}
MTCs can be either definitive or imprecise. Definitive MTCs can be further categorized into three classes: dependency, frequency, and interval. Imprecise MTCs can  be categorized into four classes: dependency, time dependency, consistency, and time-of-day.

\begin{enumerate}

\item Definitive dependency constraints capture temporal dependencies between taking medication and other regular activities. For example, from the DUG for the drug Protonix: "If you are taking the granules, take your dose \textbf{30 minutes before a meal}." 

\item Frequency constraints capture the temporal constraints regarding the suggested frequency of a medication administration, i.e., how many times a medication should be taken in a specific interval. For example, from the DUG for the drug Wellbutrin: "Take this medication by mouth, with or without food, usually \textbf{three times daily}."

\item Interval constraints capture the temporal constraints regarding the suggested interval between consecutive medication administrations. For example, again from the DUG for the drug Wellbutrin: "It is important to take your doses at least \textbf{6 hours apart} or as directed by your doctor to decrease your risk of having a seizure."

\item Imprecise dependency constraints capture inexact temporal dependencies between taking medication and other regular activities. For example, from the DUG for the drug Singulair: "Do not \textbf{take a dose before exercise} if you are already taking this medication daily for asthma or allergies. Doing so may increase the risk of side effects."

\item Time dependency constraints capture inexact temporal dependencies between taking medication and a specific time of day. For example, from the DUG for the medication Prednisone: "If you are prescribed only one dose per day, take it in the morning \textbf{before 9 AM}."

\item Consistency constraints capture the requirement to take medication consistently at a given time interval. For example, from the DUG for the medication Zocor: "Remember to take it at the \textbf{same time each day}."

\item Time-of-day constraints capture the requirement to take a medication at a certain time of a day. Take, for example, the DUG for the medication Prednisone: "If you are prescribed only one dose per day, take it \textbf{in the morning}."

\end{enumerate}

A DUG may contain multiple MTCs for a single medication. For instance, consider the following statement from the DUG of the drug Starlix: "Take this medication by mouth \textbf{1-30 minutes before each main meal}, usually \textbf{3 times daily}, or as directed by your doctor." Here the text has both a definitive dependency constraint (MTC type 1) and a frequency constraint (MTC type 2).

\subsection{A Context-free Grammar for Modeling MTCs}
A formal grammar is "context-free" if its production rules can be applied regardless of the context of a nonterminal. The taxonomy of the MTCs mentioned above motivates us to develop a context-free grammar (CFG) to model these definitive and imprecise MTCs. A CFG is a suitable solution to model MTCs as the production rule can be applied to any relevant dataset regardless of the context of the nonterminal, i.e., different types of MTCs. Our novel grammar developed and integrated in this work contains the following set of terminals. 

\begin{itemize}
    \item natural number, $n$: 1 $\mid$ 2 $\mid$ 3...
    \item activity, $act$: sleeping $\mid$ eating $\mid$ taking medication $\mid$ ...
    \item prepositions of temporal dependency, $dp$: before $\mid$ after 
    \item prepositions of interval dependency, $ip$: within $\mid$ for $\mid$ apart    
    \item prepositions of occurrence, $p$: at $\mid$ in
    \item unit of time slots, $u$: hour $\mid$ minute $\mid$ day $\mid$ week
    \item time stamp, $t$: the same time $\mid$ 9 am $\mid$ 10.30 pm $\mid$ ...
    \item time of the day, $d$: morning $\mid$ evening $\mid$ noon
\end{itemize}

Using these terminals, MTCs can be expressed using the following nonterminals.

\begin{enumerate}
\item Definitive dependency constraint: $V_1$: $n$.$u$.$dp$.$act$ (e.g., 30 minutes before eating)
\item Frequency constraint:  $V_2$: $n$ times in a $u$ (e.g., three times a day)
\item Interval constraint: $V_3$: $n$.$u$.$ip$ (e.g., 6 hours apart)
\item Imprecise dependency constraint: $V_4$: $dp$.$act$ (e.g., before meal)
\item Imprecise time dependency constraint: $V_5$: $dp$.$t$ (e.g., before 9 AM)
\item Consistency constraint: $V_6$: $p$.$t$ each $u$ (e.g., at the same time each day or at 9 am each day)
\item Time-of-day constraint: $V_7$: $p$.$d$ (e.g., in morning)
\end{enumerate}

The proposed CFG can also be used to model compound MTCs. For instance, taking a medication 2 hours before eating ($V_1$), 3 times a day ($V_2$), and 4 hours apart ($V_3$) can be expressed as: $V_i$: $V_1$.$V_2$.$V_3$. This grammar can also be extended to model negated MTCs. For instance, "do not take this medication before exercise" can be modeled as, $\lnot$ $V_4$, where $V_4$: $dp$.$act$. 


\subsection{The MTC Extraction Task}
\label{sec:extraction_task}
We define the task of \textbf{MTC extraction} as an information extraction \textbf{text-to-structure task}, in which a DUG is taken as input and a list of MTCs conforming to the proposed CFG is given as output. Using a CFG for MTC outputs blends readability and parsability. Consider this statement from the DUG for the drug Pantoprazole, which is used to treat stomach ulcers: "If you are also taking Sucralfate, take Pantoprazole at least 30 minutes before Sucralfate." The MTC contained in this statement is a definitive dependency constraint (MTC type 1), which under the CFG is labeled "30 minute before taking Sucralfate." This label is easy to understand while also conforming to the proposed CFG, i.e. $n=$ 30 $u=$ minute $dp=$ before $act=$ taking Sucralfate. Because it conforms to the CFG, the label enables potential downstream systems to model the semantics of the MTC. 

MTCs contained in free-format DUGs may not always conform to the CFG exactly. Consider the statements "take this medication at least 1 hour before any meals" and "be sure to wait at least 1 hour after taking this medication before eating." Both statements contain the definitive dependency constraint "1 hour before eating" (MTC type 1) however neither statement directly conforms to the CFG. This highlights that grammar-based decoding alone cannot accomplish the task of extracting and normalizing MTCs from these materials. Hence, we examine other methods for extracting MTCs. 

\subsection{In-Context Learning for MTC Extraction}
Guided by recent breakthroughs in clinical information extraction using LLMs \cite{agrawal2022large, dunn2022structured, torii2023task}, we explore using ICL to benchmark the MTC extraction task. ICL is a recently-introduced paradigm in few-shot sequence-to-sequence text modeling in which an LLM is asked to perform a task after being given a prompt and several examples \cite{agrawal2022large}. We choose to use GPT-3 \cite{brown2020language} in our MTC extraction experiments because it is effective in extracting both structured scientific information \cite{dunn2022structured} and medication information, such as dosage and frequency \cite{agrawal2022large}, using ICL strategies. Additionally, the number of DUGs across our three datasets is relatively small ($<1000$). Lehman et al. show that for size-constrained datasets, ICL with GPT-3 outperforms task-specific models on various clinical tasks \cite{lehman2023we}. We design two prompts for extracting MTCs from free-format textual DUG data using ICL, and a third model which utilizes customized/specialized prompts for each MTC type.

\textbf{Simple prompt:} This is a simple prompt for extracting all listed MTCs to serve as an ICL baseline. 
    
\textbf{Guided prompt:} The second is a much longer prompt, featuring elements of the labeling guide given to human annotators for annotating the FDA dataset. This prompt includes the rules of the CFG, including lists of both the terminals and the nonterminals. It also includes a list of potential activities. This is referred to as the \textit{guided} prompt.

\textbf{Specialized prompt:} We also develop prompts for extracting each of the MTC types separately. Each of the prompts contains a basic description of the MTC, as well as a heuristic for formatting the MTC correctly. This approach is referred to as the \textit{specialized} model.


\begin{figure}[htbp]
\centerline{\includegraphics[width=\columnwidth]{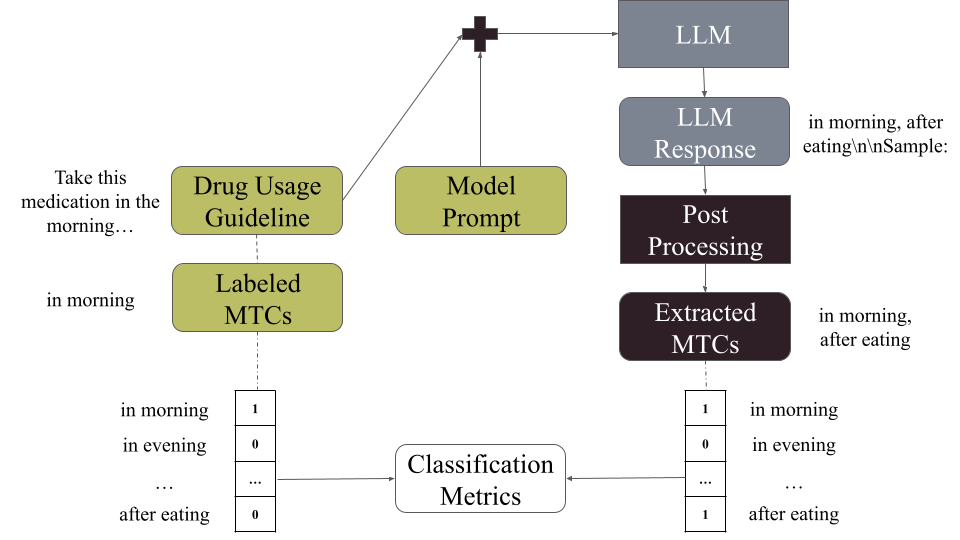}}
\caption{Overview of the In-Context Learning Text-to-Structure System}
\label{fig:system_chart}
\end{figure}

In addition to each prompt, 20 DUGs are strategically selected from the datasets and passed as few-shot examples, as in \cite{agrawal2022large}. Specifically, these 20 DUGs are selected from all three datasets such that we have a representative sample distribution for each MTC type, i.e., simple and difficult normalization examples, empty and non-empty examples, and examples of single and multiple extracted MTCs. The same 20 examples are used when testing ICL for MTC extraction across all datasets. These examples are removed from the datasets when testing to ensure no data leakage occurs. LLM output is passed through a simple post-processing module which attempts to align the outputs with the CFG. An overview of the ICL system is shown in Fig. \ref{fig:system_chart}.



%% file: sections/data.tex
\section{Data}

MTCs in DUGs can originate from doctors' suggestions, patient education materials, or guidelines for prescription medications. We utilize three datasets in this work: the FDA dataset, the Medscape dataset, and the EHR dataset \cite{openFDA, Medscape, mtsamples}. The first two datasets are derived from patient education materials for prescription medications, while the latter is sourced from prescriptions in EHRs. We use a variety of data sources to demonstrate that our MTC formalization generalizes across DUG domains. The three datasets contain a total of $N=836$ DUGs with labeled MTCs. The labeled datasets are made publicly available to enable future research in MTC extraction\footnote{\url{https://zenodo.org/record/7712934#.ZAnurj3MJD9}}. Examples from each dataset are presented, along with their labeled MTCs, in Table \ref{mtc_examples}. 

\subfile{../tables/mtc_examples}

\subsection{FDA Dataset}
The openFDA database contains drug product labels for both prescription and over-the-counter drugs submitted to the U.S. Food and Drug Administration (FDA), with text fields such as indications for use, adverse reactions, etc \cite{openFDA}. In this work we utilize the dosage and administration text field, which contains "information about the drug product's dosage and administration recommendations, including starting dose, dose range, titration regimens, and any other clinically significant information that affects dosing recommendations" \cite{openFDA}. To obtain the FDA dataset, a random sample of 600 drug labels was taken from this database. For each of the 600 drug products sampled, the dosage and administration instructions were annotated for MTCs by two annotators. Using Krippendorff's alpha coefficient for nominal data, a common measure of inter-annotator agreement for multi-label annotations \cite{krippendorff2011computing}, this annotation resulted in an agreement of 0.74, indicating good agreement. Of the 600 dosage and administration instructions, 371 contained MTCs as defined by our CFG. We refer to these drug product dosage and administration instructions as DUGs, and we refer to these 371 labeled DUGs as the FDA dataset.

\subsection{Medscape Dataset}
The Medscape dataset is sourced from 35 real prescriptions of patients with multiple chronic diseases \cite{mtsamples}, which combined include 83 unique medications. These medications treat several chronic diseases, including but not limited to diabetes mellitus (type I and type II), bipolar affective disorder, depression, hypertension, hypotension, chronic pain, morbid obesity, osteoarthritis, and obstructive sleep apnea. For each of these medications, one or more corresponding DUGs are extracted from a DUG corpus, Medscape \cite{preum2018corpus, Medscape}. From there, the MTC annotation in the DUG was a three-step process. First, three annotators annotated each sentence in each DUG for whether that sentence contained an MTC or other medical constraints, with 99.4\% agreement among all annotators as described in \cite{preum2018corpus}. Second, using a rule base, common temporal phrases were automatically assigned to these DUGs. Finally, a single annotator normalized these automatically extracted phrases to conform to the CFG. It was feasible to assign these MTCs semi-automatically because of recurring lexical patterns of MTCs in the DUG corpus. This process resulted in 121 DUGs, each annotated with one or more normalized MTCs. 

\subsection{EHR Dataset}
The EHR dataset was extracted automatically from MTSamples, a site containing a large collection of publicly-available, de-identified medical reports submitted by clinics in various medical fields, such as Gastroenterology and Pediatrics \cite{mtsamples}. These reports are submitted by many different clinicians, ensuring heterogeneity among extracted DUGs in the EHR dataset. The automatic extraction process involved searching each EHR sample for abbreviated forms of common MTCs. Healthcare professionals use medical abbreviations when writing prescriptions and medical records, some of which directly correspond to MTCs. For example, in this DUG taken from the EHR dataset "the patient has a history of lupus, currently on Plaquenil 200-mg b.i.d.", the abbreviation "b.i.d." (Latin "bis in die") means twice a day. This abbreviation maps to the frequency constraint MTC "two times a day" (MTC type 2). While there are many abbreviations in EHRs, we select 8 that map directly to MTCs. These are listed below with their matching MTCs.
\begin{enumerate}
    \item \textit{b.i.d.}: 2 times day (MTC type 2)
    \item \textit{q.d.}: 1 times day (MTC type 2)
    \item \textit{q.h.}: 1 times hour (MTC type 2)
    \item \textit{q.i.d.}: 4 times day (MTC type 2)
    \item \textit{t.i.d.}: 3 times day (MTC type 2)
    \item \textit{h.s.}: before sleep (MTC type 4)
    \item \textit{p.c.}: after eating (MTC type 4)
    \item \textit{a.c.}: before eating (MTC type 4)
\end{enumerate}
We automatically search through all the EHRs on MTSamples and extract single-sentence statements of appropriate length which include these abbreviations. Using this method, we extract 344 medical report statements and automatically assign MTC labels. 

\subsection{Data Characterization}
\label{sec:data_char}

There are 836 labeled DUGs across the three datasets; 371 from the FDA dataset, 121 from the Medscape dataset, and 344 from the EHR dataset. Combined these 836 DUGs contain 1,051 MTCs.

The use of three datasets from different sources supports the generalizability of our novel MTC taxonomy. This taxonomy can be used to identify MTCs on both the patient and provider sides since the FDA and Medscape datasets are patient-facing while the EHR dataset is clinician-facing. Statements in the FDA and Medscape datasets typically use the 2nd person perspective when discussing the patient, e.g. "to help \textit{you} remember, use it at the same time each day." Statements about patients in the EHR dataset, however, are expressed in 3rd person, e.g. "\textit{she} was finally put on Effexor 25 mg two tablets h.s."

\begin{figure}[htbp]
\centerline{\includegraphics[width=\columnwidth]{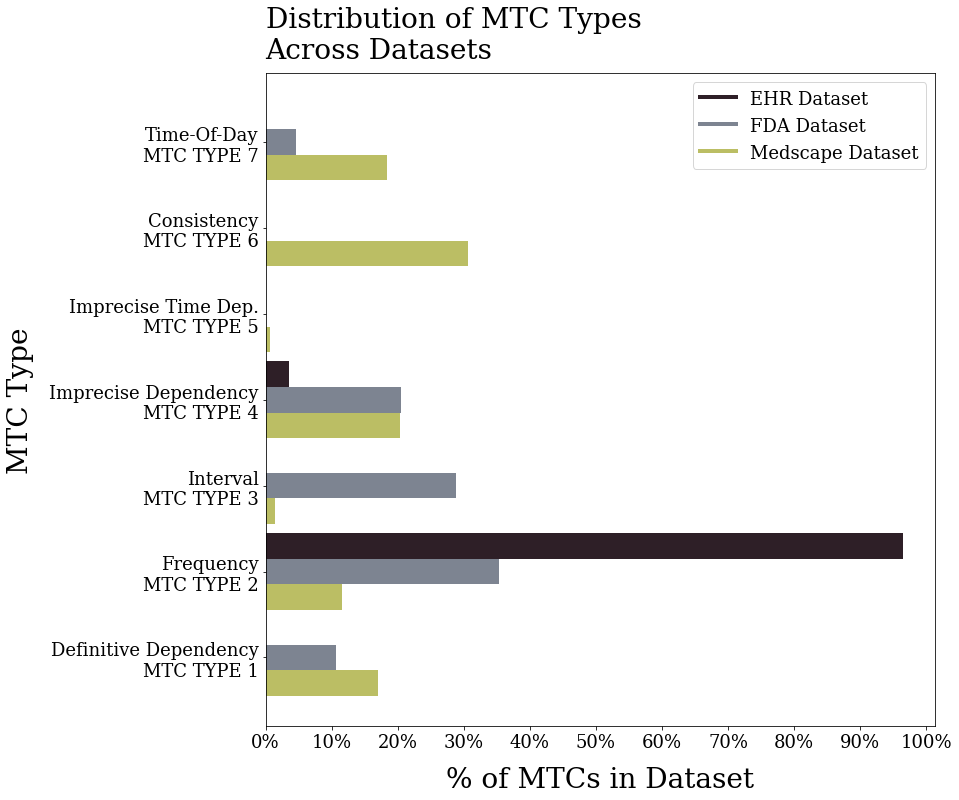}}
\caption{Distribution of MTC types across the EHR, FDA, and Medscape datasets. Along the y-axis, there are the 7 different MTC types, and the height of each bar represents the percentage of the given dataset made up of that MTC type.}
\label{fig:mtc_types_dist}
\end{figure}

While each dataset contains several MTC types, the distribution of MTC types differs across datasets, as seen in Fig. \ref{fig:mtc_types_dist}.  For instance, MTC type 6 is the most common MTC in the Medscape dataset, while it does not appear in the FDA dataset. Additionally, the EHR dataset contains almost exclusively frequency MTCs (type 2); 96.51\% of MTCs in this dataset are type 2, while the rest are type 4 MTCs. Such idiosyncrasies occur as DUGs from different sources vary with underlying medical conditions and corresponding medications/drugs. This suggests that the CFG is appropriate for multiple types of DUGs. However, MTC-type distributions may vary across DUG domains. While we explore MTCs in drug product dosage and administration labels, prescription drug labels, and de-identified medical records, MTCs may occur in other DUGs such as those found on health education websites, doctor recommendations, and elsewhere.

The ability to computationally represent MTCs is vital for downstream tasks. Hence, MTC labels provided by the annotators conform to the proposed CFG and can be represented symbolically. As an example of a downstream task that utilizes MTC extraction, consider the task of discovering whether a chronic disease patient has violated an MTC of one of their prescription medications. Based on activity patterns recognized by a human activity recognition system, a system could use MTCs extracted from the patient's prescription medication label to determine whether the patient has violated an MTC, which may lead to poor health outcomes. 

%% file: tables/mtc_examples.tex
\begin{table*}[!h]
\centering 
\caption{Examples of medical temporal constraints (MTCs) in drug usage guidelines from the FDA, Medscape, and EHR datasets}

\begin{tabular}{|c|c|c|}
\hline
\thead{Dataset} & \thead{Drug Usage Guideline} & \thead{Medical Temporal Constraints (Type)} \\
\hline
\makecell{FDA} & 
\makecell{The recommended starting dosage of donepezil hydrochloride tablets is \\ 5 mg administered once per day in the evening, just prior to retiring.}  & 
\makecell{in evening (7), 1 times day (2), before sleep (4)}  \\ 
\hline
\makecell{FDA} & 
\makecell{1-10 drops under the tongue, 3 times a day or as directed by a health \\ professional. Consult a physician for use in children under 12 years of age.}  & 
\makecell{3 times day (2)}  \\ 
\hline
\makecell{Medscape} & 
\makecell{To help you remember, use it at the same time each day.}  & 
\makecell{same time each day (6)}  \\ 
\hline
\makecell{Medscape} & 
\makecell{Do not lie down for at least 10 minutes after you have taken this drug.}  & 
\makecell{not 10 minute before sleep (1)}  \\ 
\hline
\makecell{EHR} & 
\makecell{I will initiate the sodium bicarbonate 650 mg three tablets t.i.d.}  & 
\makecell{3 times day (2)}  \\ 
\hline
\makecell{EHR} & 
\makecell{She was finally put on Effexor 25 mg two tablets h.s.}  & 
\makecell{before sleep (4)}  \\ 
\hline
\end{tabular}
\label{mtc_examples}
\end{table*}

%% file: sections/experiments.tex
\section{Experimental Setup}

\subsection{Classification Metrics}
As described in Section \ref{sec:extraction_task}, the MTC extraction task is an information extraction text-to-structure task. We choose this task structure to ensure generalizability since there are many possible MTCs according to the proposed CFG and many possible sources of MTCs. For simplicity, however, we evaluate the MTC extraction task as a multiclass multilabel classification task, with each unique extracted MTC treated as a label for a given DUG statement. We consider the union of the MTCs present in the FDA, Medscape, and EHR datasets as the label space, and include an "undefined" label for predicted MTCs that either do not conform to the CFG or do not match any MTCs in the label space. Doing so allows us to use standard multilabel classification evaluation metrics to measure model performance in the MTC extraction task. Unless stated otherwise, we henceforth report the macro average of precision, recall, and F1 \cite{scikitlearn}. 

\subsection{Validity}
In addition to standard classification evaluation metrics, we experiment using a simple heuristic for determining whether an extracted MTC is valid. We define a \textbf{valid} MTC as one which conforms to our proposed CFG, making it able to be represented computationally and thus more useful in downstream tasks. While all invalid extracted MTCs will be incorrect classifications, some incorrectly extracted MTCs will still be valid, indicating that the LLM is learning to format output according to the CFG. Hence, we report the percentage of extracted MTCs that are valid.

\subsection{Label Specifics}
In the \textit{specialized} model, since not every DUG contains an MTC of each type, some will have an empty label. We simply insert the label "NONE" for these guidelines to allow the LLM to give a non-empty response. Since there is a large prevalence of empty labels, when investigating the \textit{specialized} model specifically, we report both the positive class metrics (i.e. macro average metrics across guidelines when excluding guidelines with empty labels) and the macro average metrics across all DUGs.

We note that MTC Type 5 only occurs once across the three datasets, as seen in Fig. \ref{fig:mtc_types_dist}, and is consequently omitted from experimental results. Although MTC type 5 is discarded from further evaluation, our proposed solutions can be extended to MTC type 5 when there is relevant data.



\section{Experimental Results}
\label{sec:results}

We thoroughly examine the scope of ICL for MTC extraction. First, we compare the \textit{simple}, \textit{guided}, and \textit{specialized} ICL prompting strategies for the MTC extraction task. Next we compare ICL performance against a rule-based MTC type classification model. We then further evaluate the \textit{specialized} ICL model responses, first by dataset and then by MTC type. Finally, we explore the effectiveness of the ICL model responses to extract valid structures from text in the MTC extraction task.

\subsection{Prompting Strategies Comparison}
Macro average results on the MTC extraction task for the \textit{simple} and \textit{guided} prompts, along with the \textit{specialized} model, are displayed in Table \ref{table:in_context_results}. While the \textit{simple} and \textit{guided} prompts produce poor results overall, the \textit{specialized} model is able to competently extract MTCs with an F1 score of $0.59$. We hypothesize that extracting MTCs of each type separately, as in the \textit{specialized} model, allows the LLM to contextualize each MTC type more quickly with fewer examples.

\subfile{../tables/in_context_results}

\subsection{Rule-based Baseline Comparison}
To demonstrate the generalizability of the \textit{specialized} ICL model, we develop a simple rule-based MTC type classification model using common phrases in the Medscape dataset as guidance. As extracting specific MTCs using a simple set of search rules would be difficult, we instead attempt only to identify which MTC types (1-7) occur in a given DUG. We use only the Medscape dataset when developing the rule base, as it has the greatest variety of MTCs, then use the same rule base to identify MTC types across all three datasets. We see in Table \ref{table:rulebase_comparison} that even when only attempting to identify MTC types in a DUG, a much simpler task than MTC extraction, a rule base developed for the Medscape dataset fails to generalize to either the FDA dataset or the EHR dataset. In comparison, the \textit{specialized} model generalizes across all 3 datasets in the MTC extraction task. This demonstrates the generalizability of using ICL for the MTC extraction task.

\subfile{../tables/rulebase_comparison_results}
\subsection{Specialized ICL Results By Dataset}
In Table \ref{table:icl_by_dataset} we see the results of the \textit{specialized} model across each of the three datasets\footnote{\label{refnote}As explained in Section \ref{sec:data_char}, only one imprecise time-of-day dependency MTC (type 5) occurs across the EHR, FDA, and Medscape datasets. Hence, we do not attempt to extract this MTC type.}. The \textit{specialized} model performs the best across the EHR dataset, with a macro average F1 score of $0.65$. The EHR dataset presents possibly the easiest of the three MTC extraction tasks because all labeled MTCs are mapped one-to-one with medical abbreviations, and there are only two MTC types present in the dataset. 

\subfile{../tables/icl_by_dataset}

\subsection{Specialized ICL Results By MTC Type}
In Table \ref{table:in_context_by_mtc} we see the results of the \textit{specialized} model by MTC type\footref{refnote}. While the model is able to accurately extract MTCs of most types, it performs best on interval constraints (MTC type 3) with a positive class macro average F1 score of $0.72$. The most difficult MTC type for the model to extract is the consistency MTC (type 6), with a positive class macro average F1 score of $0.33$. We see that the \textit{specialized} model frequently hallucinates consistency MTCs and discusses other potential sources of error in Section \ref{section:error_analysis}.

\subfile{../tables/in_context_by_mtc}

\subsection{Validity}
Finally, we explore the ability of the ICL models to produce parsable outputs. We see in Table \ref{table:validity} that the \textit{specialized} model is far more competent at producing parsable output which conforms to the CFG with minimal post-processing, with a $0.99$ proportion of valid outputs compared to $0.29$ and $0.37$ in the \textit{simple} and \textit{guided} models, respectively.

\subfile{../tables/validity}

\section{Error Analysis}
\label{section:error_analysis}

To investigate model strengths and weaknesses, we sample 60 errors made by the \textit{specialized} model, 10 of each MTC type. A single human annotator then categorizes each model error, providing one possible reason for each failed MTC extraction. The three most frequent error categories in this sample are hallucinations, semantic overlap, and nonvalidity. The sample distribution of these error categorizations across MTC types is provided in Table \ref{table:error_analysis}. Examples of each of these common error types are given in Table \ref{error_examples}. We now describe each of the three common error types. 

\subfile{../tables/error_analysis}

\subfile{../tables/error_examples}

The most common error type is \textbf{hallucination}, an error common in LLMs such as GPT-3 \cite{ji2022survey}. This occurs when the model outputs an MTC or a list of MTCs that are valid, but not found in the text sample. 43\% of all labeled model errors are due to hallucinations. A common cause of hallucination occurs in activity selection. Definitive and imprecise dependency constraints (types 1 and 4, respectively) occur when the medication intake activity is temporally dependent on another patient's activity. Human annotators were instructed to normalize activities when labeling MTCs. For example, phrases like "before bedtime" and "before sleeping" were both normalized to "before sleep." The \textit{specialized} model occasionally either hallucinates an activity, fails to normalize an activity or both. Hallucinations were especially common in consistency (type 6) and time-of-day (type 7) MTCs, accounting for 70\% of these errors in the categorized sample. An example of a consistency (type 6) MTC hallucination is given in the second row of Table \ref{error_examples}. Hallucinations seem to be a primary reason for poor model performance when extracting consistency MTCs (type 6) specifically, as the positive class macro average F1 was quite low ($0.33$) but overall performance was much higher ($0.63$ macro average F1). Reducing hallucinations in LLMs is an active area of research that could lead to better results on the MTC extraction task \cite{sun2022contrastive}.

Another common error is \textbf{nonvalidity}, in which LLM model output is unable to be parsed according to the CFG, after minimal post-processing. Take the first DUG in Table \ref{error_examples}, from the Medscape dataset. While the \textit{specialized} model output "2 times day OR 3 times day" is not semantically incorrect, the inclusion of the "OR" makes this output nonvalid according to the CFG. Nonvalidity was the primary error type of 15\% of the categorized model errors.

The final common error type among the labeled sample errors is \textbf{semantic overlap}. Under the proposed CFG, certain MTCs can be accurately represented by different MTC types. Consider, for example, the last DUG in Table \ref{error_examples}, from the FDA dataset. While the direction to take the medication "at morning and at noon" could potentially imply the "6 hours apart" interval constraint, as extracted by the \textit{specialized} model, this was instead labeled as the two semantically-related time-of-day MTCs "in the morning" and "at noon". Semantically-overlapping errors primarily occurred in 15\% of the categorized sample errors.

%% file: tables/in_context_results.tex
\begin{table}[htbp]
\begin{center}
\caption{MTC Extraction Results. We report macro average classification metrics recall, precision, and F1 score.}
\begin{tabular}{|l|r|r|r|r|r|}
\hline
\textbf{Model} &\textbf{Recall} &\textbf{Precision} &\textbf{F1} \\
\hline
\textit{Simple} &0.12 &0.14 &0.12 \\
\hline
\textit{Guided} &0.15 &0.21 &0.17 \\
\hline
\textit{Specialized} &\textbf{0.57} &\textbf{0.70} &\textbf{0.59} \\
\hline
\end{tabular}
\label{table:in_context_results}
\end{center}
\end{table}

%% file: tables/rulebase_comparison_results.tex
\begin{table}[htbp]
\begin{center}
\caption{Comparison of rule-based model and \textit{specialized} Model by Dataset. The rule-based model predicts which MTC types occur in each drug usage guideline, whereas the \textit{specialized} model is used for the MTC extraction task. We report macro average F1 across each dataset for both tasks.}
\begin{tabular}{|l|r|r|r|}
\hline
\thead{Model} &\thead{Medscape \\ Dataset F1} &\thead{FDA \\ Dataset F1} &\thead{EHR \\ Dataset F1} \\
\hline
\makecell{Rule-Based Model} &0.63	&0.43	&0.01 \\
\hline
\makecell{\textit{Specialized}} &0.59	&0.61	&0.65 \\
\hline
\end{tabular}
\label{table:rulebase_comparison}
\end{center}
\end{table}

%% file: tables/icl_by_dataset.tex
\begin{table}[htbp]
\begin{center}
\caption{\textit{Specialized} Model Results by Dataset. For example, the \textit{Specialized} model is able to extract interval MTCs in the Medscape and FDA datasets with macro F1 scores of $0.50$ and $0.70$, respectively, whereas no interval MTCs are labeled in the EHR dataset.}
\begin{tabular}{|l|r|r|r|}
\hline
\thead{MTC Type} &\thead{Medscape \\ Dataset F1} &\thead{FDA  \\ Dataset F1} &\thead{EHR \\ Dataset F1} \\
\hline
\makecell{Definitive \\ Dependency (1)} &0.65	&0.80	&-- \\
\hline
\makecell{Frequency (2)} &0.63	&0.57	&0.79 \\
\hline
\makecell{Interval (3)} &0.50	&0.70	&-- \\
\hline
\makecell{Imprecise \\ Dependency (4)} &0.45	&0.38	&0.38 \\
\hline
\makecell{Consistency (6)} &0.63	&--	&-- \\
\hline
\makecell{Time-of-Day (7)} &0.53	&0.50	&-- \\
\hline
\makecell{\textit{Overall}} &\textit{0.59}	&\textit{0.61}	&\textit{0.65} \\
\hline
\end{tabular}
\label{table:icl_by_dataset}
\end{center}
\end{table}

%% file: tables/in_context_by_mtc.tex
\begin{table}[htbp]
\begin{center}
\caption{\textit{Specialized} Model Results by MTC Type. We report macro average metrics for both the positive class and overall. We report both the positive class F1 score (i.e. macro average metrics across guidelines when excluding guidelines with empty labels) since there is a large prevalence of empty labels.}
\begin{tabular}{|l|r|r|r|r|r|r|r|}
\hline
\thead{MTC \\ Type} &\thead{Positive \\ Recall} &\thead{Positive \\ Precision} &\thead{Positive \\ F1} &\thead{Recall} &\thead{Precision} &\thead{F1} \\
\hline
\makecell{1} &0.67 &0.65	&0.65	&0.67	&0.70	&0.67 \\
\hline
\makecell{2} &0.64	&0.57	&0.60	&0.65	&0.66	&0.64	 \\
\hline
\makecell{3} &0.71	&0.74	&0.72	&0.64	&0.81	&0.68	 \\
\hline
\makecell{4} &0.46	&0.46	&0.42	&0.36	&0.49	&0.38	 \\
\hline
\makecell{6} &0.33	&0.32	&0.33	&0.61	&0.65	&0.63	 \\
\hline
\makecell{7} &0.49	&0.50	&0.49	&0.37	&0.66	&0.43	\\
\hline
\end{tabular}
\label{table:in_context_by_mtc}
\end{center}
\end{table}

%% file: tables/validity.tex
\begin{table}[htbp]
\begin{center}
\caption{Validity of Extracted MTCs by Model Type}
\begin{tabular}{|l|r|r|r|r|r|}
\hline
\textbf{Model} &\textbf{Validity} \\
\hline
\textit{Simple} &0.29 \\
\hline
\textit{Guided} &0.37 \\
\hline
\textit{Specialized} &\textbf{0.99} \\
\hline
\end{tabular}
\label{table:validity}
\end{center}
\end{table}

%% file: tables/error_analysis.tex
\begin{table}[htbp]
\begin{center}
\caption{\textit{Specialized} Model Error Counts by MTC Type. For example, of the 10 labeled consistency MTC (type 6) extraction errors, 7 are hallucinations. The other 3 have undetermined error sources.}
\begin{tabular}{|l|r|r|r|r|}
\hline
\thead{MTC Type} &\thead{Hallucination} &\thead{Semantic \\ Overlap} &\thead{Nonvalidity} & \thead{Other} \\
\hline
\makecell{Definitive \\ Dependency (1)} &3	&2	&1	&4 \\
\hline
\makecell{Frequency (2)} &0	&0	&4	&6 \\
\hline
\makecell{Interval (3)} &5	&1	&3	&1 \\
\hline
\makecell{Imprecise \\ Dependency (4)} &4	&4	&0	&2 \\
\hline
\makecell{Consistency (6)} &7	&0	&0	&3 \\
\hline
\makecell{Time-of-Day (7)} &7	&2	&1	&0 \\
\hline
\end{tabular}
\label{table:error_analysis}
\end{center}
\end{table}

%% file: tables/error_examples.tex
\begin{table*}[!h]
\centering 
\caption{Examples of In-Context Learning Errors. The second error, for example, is a hallucination of the consistency MTC "same time each day" (type 6) given by the \textit{specialized} model when attempting to extract consistency MTCs.}
\begin{tabular}{|c|c|c|c|}
\hline
\thead{Drug Usage Guideline} & \thead{MTCs (Type Expected)} & \thead{Model Output} & \thead{Error Type} \\
\hline
\makecell{Take this medication by mouth as directed \\ by your doctor, usually 2 or 3 times daily.} & 
\makecell{3 times day (2)}  & 
\makecell{2 times day OR 3 times day} &
\makecell{Nonvalidity} \\ 
\hline
\makecell{One tablet daily or as directed by a physician.} & 
\makecell{NONE (6)}  & 
\makecell{same time each day} &
\makecell{Hallucination} \\ 
\hline
\makecell{Your doctor may direct you to take it in the \\ morning and at noon.} & 
\makecell{NONE (3)}  & 
\makecell{12 hour apart} &
\makecell{Semantic Overlap} \\ 
\hline
\end{tabular}
\label{error_examples}
\end{table*}

%% file: sections/relatedworks.tex
\section{Related Works}

\subsection{Medical Information Extraction Tasks and Context-Free Grammars}

While our work is the first to formalize the MTC extraction task, the broader field of medical information extraction is a vibrant area of research. Most related to the MTC extraction task, much prior work has been done in modeling and extracting temporal and medication information in medical and health texts. A common task is extracting temporal relations between clinical events, such as problems and treatments, in discharge summaries \cite{sun2013evaluating, alfattni2020extraction}. Another is identifying medication information such as drug names, strengths, and routes in electronic medical records \cite{xu2010medex, wei2020study}. MTC extraction is related yet novel that is more patient-centric in that it focuses on extracting temporal constraints placed on health-related activities found in DUGs.

Our work additionally focuses on modeling extracted MTCs using context-free grammar. The modeling of temporal phenomena in medical text using CFGs has been leveraged by Hao et al. \cite{pan2020temporal}, who introduce a model to leverage CFG to extract temporal expressions in clinical texts. In a similar work, Viani et al. utilize a CFG to parse mental health records and extract the duration of untreated psychosis \cite{viani2020temporal}. Our work is the first to use a CFG to define and extract MTCs in DUGs, and we are the first to experiment with ICL for this task.

\subsection{In-Context Learning for Medical Information Extraction}

Agrawal, Hegselmann, Lang, Kim, and Sontag have shown that LLMs are able to extract clinical information from the medical text in both the few-shot and zero-shot settings \cite{agrawal2022large}. Specifically, they show that given inputs of clinical discharge summaries or medical abstracts, along with guided prompts, GPT-3 \cite{brown2020language} is able to competently perform many medical information extraction tasks such as clinical sense disambiguation, biomedical evidence extraction, and medication extraction. We experiment with similar strategies to benchmark the MTC extraction and normalization task. Related works that utilize ICL for structured scientific information extraction include \cite{dunn2022structured}, which extracts entities and entity relationships from scientific documents into JSON format, and \cite{torii2023task}, which formulates the task of extracting social determinants of health from clinical narratives.

%% file: sections/conclusion.tex
\section{Conclusion}

In this work, we have developed a novel taxonomy of potential MTCs and a novel CFG based model to computationally represent MTCs found in unstructured DUGs. We present and release three new datasets containing $N=836$ DUGs with labeled normalized MTCs. Finally, guided by recent work in ICL for medical information extraction, we develop and explore an ICL solution for the MTC extraction task, achieving an average F1 score of $0.62$ across all datasets. Patient-in-the-loop systems that utilize MTC extraction will have computational representations of patient constraints to guide patient activity, promote medication adherence, and lead to better health outcomes. The taxonomy and CFG of MTCs, dataset of extracted MTCs, and ICL exploration presented in this work will advance patient-centric healthcare applications for treatment adherence.